\documentclass{vgtc}                          

\ifpdf
  \pdfoutput=1\relax                   
  \usepackage{graphicx}                
  \DeclareGraphicsExtensions{.pdf,.png,.jpg,.jpeg} 
\else
  \usepackage{graphicx}                
  \DeclareGraphicsExtensions{.eps}     
\fi%

\graphicspath{{figures/}{pictures/}{images/}{./}} 

\usepackage{microtype}                 
\PassOptionsToPackage{warn}{textcomp}  
\usepackage{textcomp}                  
\usepackage{mathptmx}                  
\usepackage{times}                     
\usepackage{cite}                      
\usepackage{tabu}                      
\usepackage{booktabs}                  
\usepackage{xcolor}
\usepackage{enumitem}
\usepackage{amsmath}

\vgtccategory{Research}

\vgtcinsertpkg

\title{360 Depth Estimation in the Wild -\\ the Depth360 Dataset and the SegFuse Network}

\author{Qi Feng\thanks{e-mail: fengqi@ruri.waseda.jp}\\ %
        \scriptsize Waseda University %
\and Hubert P. H. Shum\thanks{e-mail: hubert.shum@durham.ac.uk}\\ %
     \scriptsize Durham University %
\and Shigeo Morishima\thanks{e-mail: shigeo@waseda.jp}\\ %
     \parbox{1.4in}{\scriptsize \centering Waseda Research Institute \\ for Science and Engineering}}

\abstract{Single-view depth estimation from omnidirectional images has gained popularity with its wide range of applications such as autonomous driving and scene reconstruction. Although data-driven learning-based methods demonstrate significant potential in this field, scarce training data and ineffective 360 estimation algorithms are still two key limitations hindering accurate estimation across diverse domains.
In this work, we first establish a large-scale dataset with varied settings called Depth360 to tackle the training data problem. This is achieved by exploring the use of a plenteous source of data, 360 videos from the internet, using a test-time training method that leverages unique information in each omnidirectional sequence. With novel geometric and temporal constraints, our method generates consistent and convincing depth samples to facilitate single-view estimation. 
We then propose an end-to-end two-branch multi-task learning network, SegFuse, that mimics the human eye to effectively learn from the dataset and estimate high-quality depth maps from diverse monocular RGB images. With a peripheral branch that uses equirectangular projection for depth estimation and a foveal branch that uses cubemap projection for semantic segmentation, our method predicts consistent global depth while maintaining sharp details at local regions. Experimental results show favorable performance against the state-of-the-art methods.%
} 

\CCScatlist{
  \CCScatTwelve{Computing methodologies}{Computer graphics}{Image manipulation}{Image-based rendering};
  \CCScatTwelve{Computing methodologies}{Artificial intelligence}{Computer vision}{Reconstruction}
}

\begin{document}

\firstsection{Introduction}

\maketitle

\begin{figure}[t]
\centering
  \includegraphics[width=0.48\textwidth,keepaspectratio]{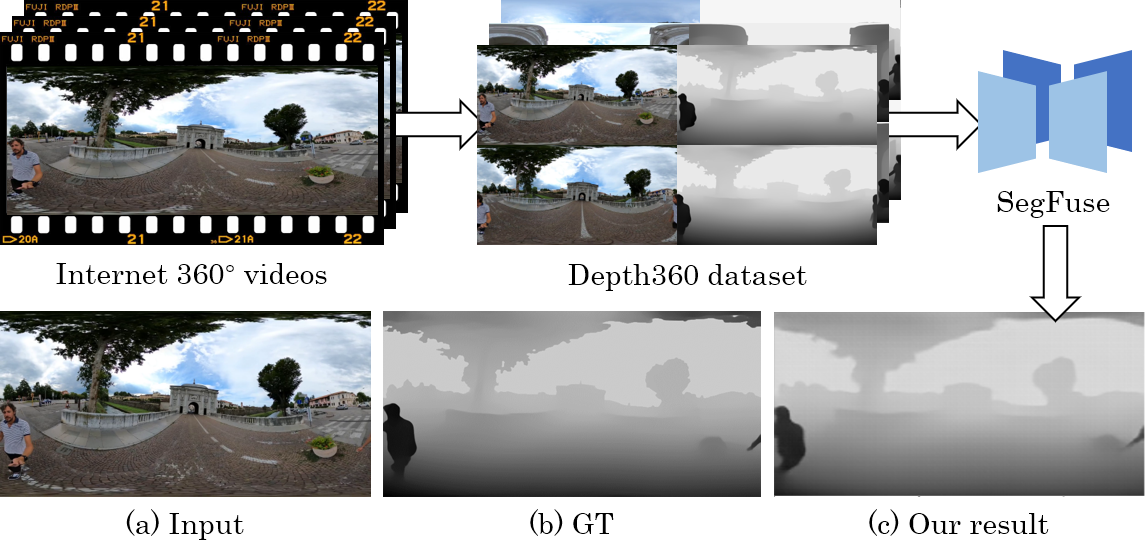}
  \vspace{-18pt} 
  \setlength{\belowcaptionskip}{-5pt}
  \caption{We present a method for generating large amounts of color/depth training data from abundant internet 360 videos. After creating a large-scale general omnidirectional dataset, Depth360, we propose an end-to-end two-branch multitasking network, SegFuse to learn single-view depth estimation from it. Our method shows dense, consistent and detailed predictions.}
  \label{fig:zero}
\end{figure}

Visual reasoning in the context of omnidirectional images has gained increasing popularity in both academic and industrial communities during the past few years. By providing rich information of the environment with large field-of-view (FOV), predicting dense depth maps from a single 360 image shows wide applicability and facilitates applications that require accurate understandings of the context, such as scene reconstruction \cite{serrano2019motion} and autonomous navigation \cite{gaspar2000vision}. However, inferring depth from a monocular image is a challenging and ill-posed problem due to uncontrolled extrinsic, ambiguous scales, and varied settings. Recently, data-driven deep learning methods \cite{laina2016deeper} have presented significant potential in this field.

Despite learning-based methods having been extensively studied within the context of perspective images, omnidirectional format presents challenges in both aspects: data preparation and depth estimation algorithm. 
On the one hand, large-scale 360 training data is difficult to collect. For synthesis-based methods, the cost to create large-scale models that resemble real-world ones with abundant settings is excessively high \cite{zhang2017physically}, and the diversity gap between synthetic samples and real data leads to less accurate results \cite{zhu2021spatially}. For capturing-based methods, using dual 360 cameras for stereo-capturing will introduce mutual occlusion. Specialized scanning devices (e.g. Matterport \cite{chang2017matterport3d}) produce dense datasets but are limited to indoor use due to their working principle. Depth maps produced with laser scanners (such as LIDAR \cite{ma2019self}) suffer from self-occlusion albeit being the main source for outdoor settings. Most datasets are only captured under specific scenarios (e.g. atop a driving car \cite{chen2018lidar}).
On the other hand, existing learning-based approaches cannot effectively take advantage of 360 image datasets. The majority of depth estimation methods \cite{laina2016deeper} are designed for perspective cameras with narrower FOV. Due to the spherical nature of the content, projecting to a 2D image introduces irregular distortions and thus hinders effective learning \cite{zioulis2019spherical}. Even though there are a few methods \cite{zioulis2018omnidepth} \cite{wang2020bifuse} proposed with distortions in mind, they only focus on indoor settings due to the unavailability of outdoor datasets. As a result, they show sub-optimal performance under general cases.

In this paper, we first tackle the problem of limited datasets by exploring the use of the plenteous source of data: 360 videos from the internet that are captured with a moving hand-held omnidirectional camera. We propose a test-time training method that utilizes a learning-based prior to synthesizing plausible depth maps for each consistent 360-degree video. 
By leveraging the rich information that is only presented in omnidirectional formats, we propose to use the output of structure-from-motion (SfM) and multi-view stereo (MVS) methods to calculate a novel geometric consistency based on a geometric spherical disparity model. We also propose to use optical flow \cite{ilg2017flownet} to encourage temporal consistency and establish multiple constraints for each pixel that ensure a convincing output. With established constraints, we fine-tune a pre-trained model by updating the parameters according to the calculated geometric and temporal losses to produce more consistent output for a particular sequence. During dataset creation, our test-time training method takes preprocessed video sequences as input and generates geometrically and temporally consistent dense depth map for each frame. To our knowledge, our large-scale dataset, Depth360, is the first to use internet omnidirectional videos for achieving monocular depth estimation from single 360 images. To benchmark the accuracy of data generation, we propose using rendering-based methods to further generate a photorealistic synthetic dataset, SynDepth360. With the unlimited training data with diverse conditions, we seek to learn depth estimation with high accuracy and generalization.

We then propose an end-to-end neural network architecture, SegFuse, to learn the single-view depth estimation of omnidirectional images that generalize well with a wide range of settings by mimicking the human eye. While videos usually provide more cues for depth calculation and facilitate dataset creation, lengthy optimization for individual scenes does not achieve as good generalization and practicality compared to single-view depth estimation. We believe that compared to indoor depth maps with more uniform distributions and relatively universal ranges, more challenging variations of outdoor images, i.e. unsymmetrical depth distributions (sky and ground) and distinct depth ranges between different scenes, lead to ineffectively learning processes and generalization for existing methods. To cope with such problems, we propose a multi-task learning framework that adopts a bi-projection fusion scheme: a peripheral branch that uses equirectangular projection for depth estimation and a foveal branch that uses cubemap projection for semantic segmentation. While equirectangular projection can provide consistent global context, cubemap projection gives more local details with a narrower FOV. With peripheral vision to perceive the depth of the scene and foveal vision to distinguish between different objects, our method can successfully learn a smooth global depth while maintaining details at local regions. Compared to the method \cite{wang2020bifuse} with a similar structure, SegFuse uses multi-task learning to exploit semantics in complex depth distributions, and achieve significantly improved performance in outdoor settings.

By applying the generated training data with diverse conditions to multiple state-of-the-art learning-based omnidirectional depth estimation methods, our experimental results show that our method outperforms existing methods with more consistent global results and sharper local estimations. 

To summarize, our contributions are as follows:
\begin{enumerate}[noitemsep,topsep=0pt]
    \vspace{0.1cm} \item To solve the unavailability of a general omnidirectional dataset with dense depth maps, we are first to propose to utilize omnidirectional video in the wild to generate a large-scale dataset, Depth360. By exploiting unique temporal and geometric consistencies of 360 videos with a spherical disparity model, we use test-time training to generate convincing depth maps.
    \vspace{0.1cm} \item We propose an end-to-end two-branch multi-task architecture called SegFuse that estimates depth from a single-view 360 image input by mimicking the human eye. The peripheral branch regresses global depth estimation while the foveal branch estimates local semantic segmentation. By fusing the global context and local details, our design ensures a sharp and consistent depth prediction under challenging cases.
    \vspace{0.1cm} \item To validate the accuracy of the proposed dataset and evaluate the effectiveness of our multitasking method, we perform an extensive evaluation against state-of-the-art omnidirectional datasets and methods and present a better quantitative and qualitative performance.
    \vspace{0.1cm} \item To encourage future research, the datasets, source codes, and models are made available to the community at:\\
    \textcolor{blue}{\url{https://github.com/HAL-lucination/segfuse}}
\end{enumerate}

\begin{figure*}[ht]
\centering
  \includegraphics[width=0.75\textwidth,keepaspectratio]{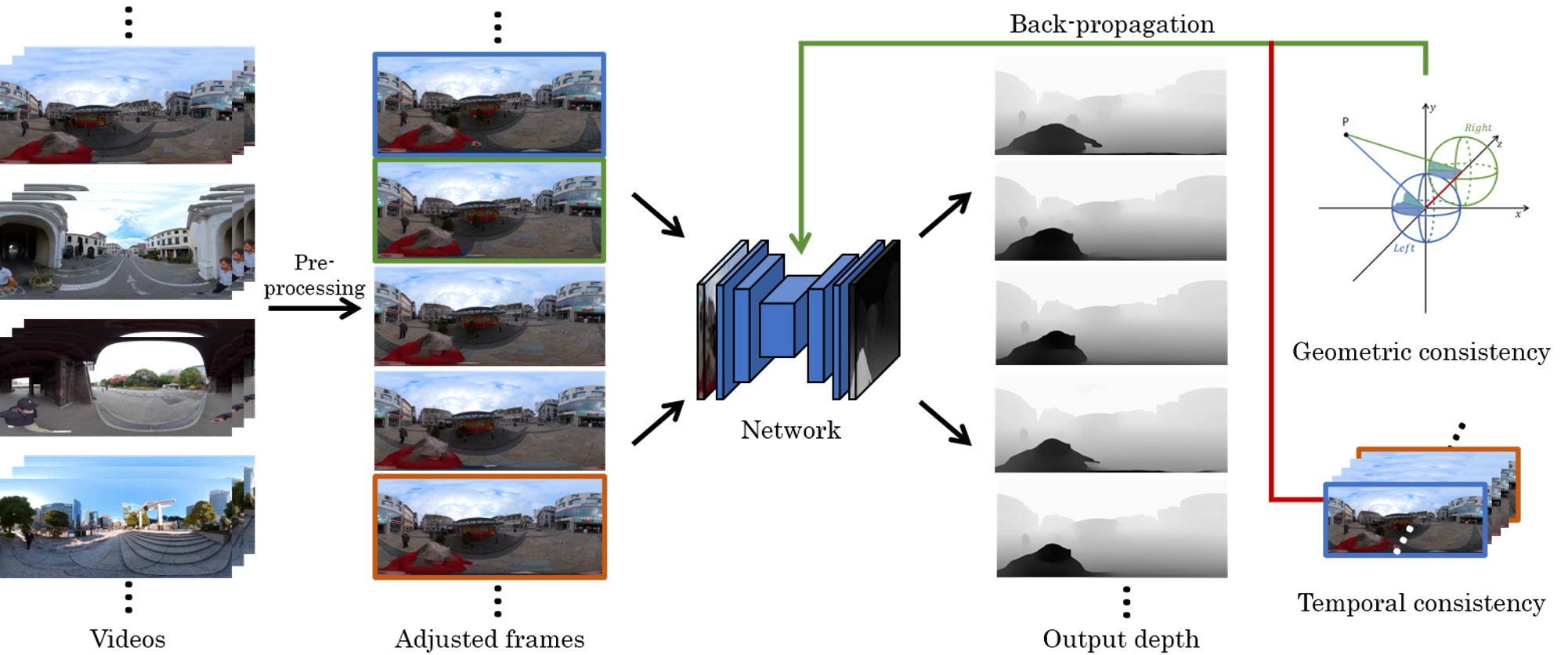}
    \setlength{\belowcaptionskip}{-15pt}
  \caption{The overview of our dataset generation method. With monocular videos as input, we samples successive frames from a single sequence and adjust the frames spatially with baselines acquired with SfM and MVS methods. Geometric and temporal constraints of this sequence are then established using a geometric spherical disparity model and a 360-aware optical flow algorithm. By fine-tuning a learning-based prior with computed losses through back-propagation during test time, we can generate consistent depth output that satisfies the constraints of the corresponding sequence.}
  \label{fig:one}
\end{figure*}

\section{Related Work}

\subsection{Data in Single-view Depth Estimation}
One of the major issues in learning-based single-view depth estimation is the unavailability of data. For perspective images, most supervised depth-estimation methods are trained on a few standard datasets (e.g. NYU \cite{silberman2012indoor}) due to the difficulty of acquiring ground truth depth maps. Capturing-based methods often utilize RGB-D sensors and laser scanning (e.g. LIDAR \cite{chen2018lidar}). To improve data availability and ease of acquisition, several efforts have been made. Godard et al. \cite{godard2017unsupervised} use multiple views of a scene as a supervisory signal, but these approaches usually require two input images at test time \cite{luo2018single}. Mayer et al. \cite{mayer2016large} uses a synthetic dataset, but the domain gap results in sub-optimal performance in real-world scenarios and requires further domain adaptation \cite{atapour2018real}. Using internet images \cite{li2018megadepth} and videos \cite{jafarian2021learning} to calculate pseudo ground truth with structure-from-motion and multi-view stereo shows great performance but is only explored in perspective context.

When it comes to omnidirectional depth maps, not only capturing-based methods are greatly limited, but also the existing perspective-based approaches are less effective, resulting in the scarcity of outdoor datasets. Existing omnidirectional sensors with customized arrays suffer from strong self-occlusions, leading to missing or sparse information at the bottom of the sphere. Using multiple monocular cameras as a stereo setup (i.e., 3D VR cameras) to calculate disparity is also problematic due to mutual occlusion \cite{richardt2013megastereo}. Using domain adaptation for synthetic data requires both large-scale 3D models with great variations and corresponding similar 360-degree color ground truth. Most concurrent works \cite{zioulis2018omnidepth} \cite{wang2020bifuse} either use synthetic datasets (i.e., PanoSunCG \cite{wang2018self}) or 3D scanned datasets (i.e., Matterport3D \cite{chang2017matterport3d}, Stanford 2D-3D \cite{armeni2017joint}, Pano3D \cite{albanis2021pano3d}). The former is generated with 3D models and a virtual omnidirectional camera without domain adaptation, and the latter ones are captured with specialized equipment and post-processed. Both suffer from no dynamic foregrounds, further limiting their usefulness in real-world scenarios \cite{cc0dcd4b355d4d8091f68a31ac903a8b}. Zhu et al. \cite{zhu2021spatially} propose to use physics-based rendering to generate synthetic outdoor panoramas, but the diversity gap between synthetic samples and real data leads to less accurate results. Therefore, taking advantage of an increasing number of shared online omnidirectional videos, we propose a pipeline to utilize rich information in the wild to generate a large-scale dataset.

\subsection{Single-view Depth Estimation for Perspective Images}
Predicting depth from monocular color images is an important task in understanding 3D scene geometries \cite{yang2019dula}. An accurate estimation can benefit various applications such as autonomous driving \cite{chen2018lidar} and graphics rendering \cite{luo2020consistent}. Traditional methods of monocular depth estimation heavily rely on probabilistic graphical models with hand-crafted local features and constraints (e.g. MRF) \cite{saxena2005learning}. With the advances in deep learning algorithms, recent learning-based approaches \cite{eigen2015predicting} \cite{godard2019digging} \cite{miangoleh2021boosting} show significant improvements in accuracy. 

A standard approach of learning an implicit relation between color and depth is to train models with collected RGB images and ground truth depth maps. Eigen et al. \cite{eigen2015predicting} propose multi-scale networks to refine coarse depth with local details. This two-scale strategy is further refined to predict high-resolution depth \cite{miangoleh2021boosting}. A fully convolutional architecture with a novel up-projection module proposed by \cite{laina2016deeper} improves the output accuracy. Cao et al. \cite{cao2017estimating} propose to solve depth regression in a classification fashion. Another direction for improving the output quality is to combine graphical models with the use of CNNs, such as incorporating conditional random fields in the form of a loss function into the depth estimation task \cite{liu2015deep}. However, when directly applying perspective models to 360 images, an inferior performance is observed due to the lack of global consistency and incorrectly modeling the projection’s distortion \cite{zioulis2018omnidepth}. 

\subsection{Single-view Depth Estimation for Omnidirectional Images}
As omnidirectional cameras have become more efficient and accessible, the interest in 360-degree media has surged on the internet owing to novel applications such as virtual reality \cite{bertel2020omniphotos} \cite{shimamura2020audio} and mixed reality \cite{rhee2017mr360}. For single-view depth estimation, while a large body of research exists for perspective images, scarce work has been done to address this problem for spherical images. The most apparent issue is the distortion introduced when projecting the 3D spherical information onto the 2D plane. Although rotation equivariant CNNs \cite{cohen2018spherical} and graph-based learning \cite{khasanova2017graph} with spherical cross-correlation directly learn from 3D spherical signals, such equivariant architectures define convolution in the spectral domain and provide a lower network capacity, hindering applicability in generative tasks such as monocular depth estimation. To apply deep learning approaches to omnidirectional content, most approaches are proposed using two projection formats, cubemap and equirectangular projections. 

While cubemap projects spherical signals onto 6 faces of a cube, and thus enables directly feeding non-distorted images into a CNN, the discontinuity along boundaries is problematic when trying to merge results back into a spherical image. A common solution is using cube padding \cite{cheng2018cube} to aid the network merging estimations for each face into a full omnidirectional output. This method is effective when applied to single-view depth estimation for indoor scenes \cite{wang2020bifuse} with a relatively uniform depth distribution and other tasks such as stylization \cite{ruder2018artistic} and classification \cite{coors2018spherenet}. However, these methods are less effective when each face has wildly changing depth ranges in outdoor scenarios \cite{vasiljevic2019diode}. Since each face only includes very limited information of a local region, dramatically different appearance and the ambiguity of depth scales usually result in distinct estimations, limiting the scalability of such approaches. Recent works using diverse division schemes show improved predictions for indoor samples \cite{sun2021hohonet}. However, slice-based methods that exploit relationships of vertical patches \cite{pintore2021slicenet} also report discontinuities for outdoor cases.

To make the network efficient and directly aware of the distortion in omnidirectional images, work resorted to using equirectangular projection with distorted filters \cite{zioulis2018omnidepth} and dilations \cite{fernandez2020corners}. However, the effectiveness of these methods is limited. As the layers deepen, non-linearly distributed information across an equirectangular image got lost (e.g. consistency across the sphere). Although this problem is alleviated by a kernel transformer \cite{su2019kernel} that uses parameterized functions to preserve cross-channel interactions, the model size is still limited. While using equirectangular projection can generate more consistent global prediction due to its wider FOV, small regions with a steep local gradient when regressing the global gradient are harder to learn \cite{feng2020foreground}. Wang et al. \cite{wang2020bifuse} and Jiang et al. \cite{jiang2021unifuse} use a fusion scheme that combines the depth maps estimated with equirectangular and cubemap projections for sharper depth estimation. Although it presents improved accuracy for indoor settings, the disadvantage of limited scalability remains \cite{pintore2021slicenet}. Instead, we purpose an architecture that fuses a cubemap branch for semantic segmentation with an equirectangular branch for depth estimation. Considering that regressed depth maps for different faces are hard to balance when training with outdoor samples, semantic segmentation can serve to inform the global depth estimation of the local details without the problem of balancing scales between each local view.
\section{The Depth360 Dataset}

\begin{figure}[htb]
\centering
  \includegraphics[width=0.48\textwidth,keepaspectratio]{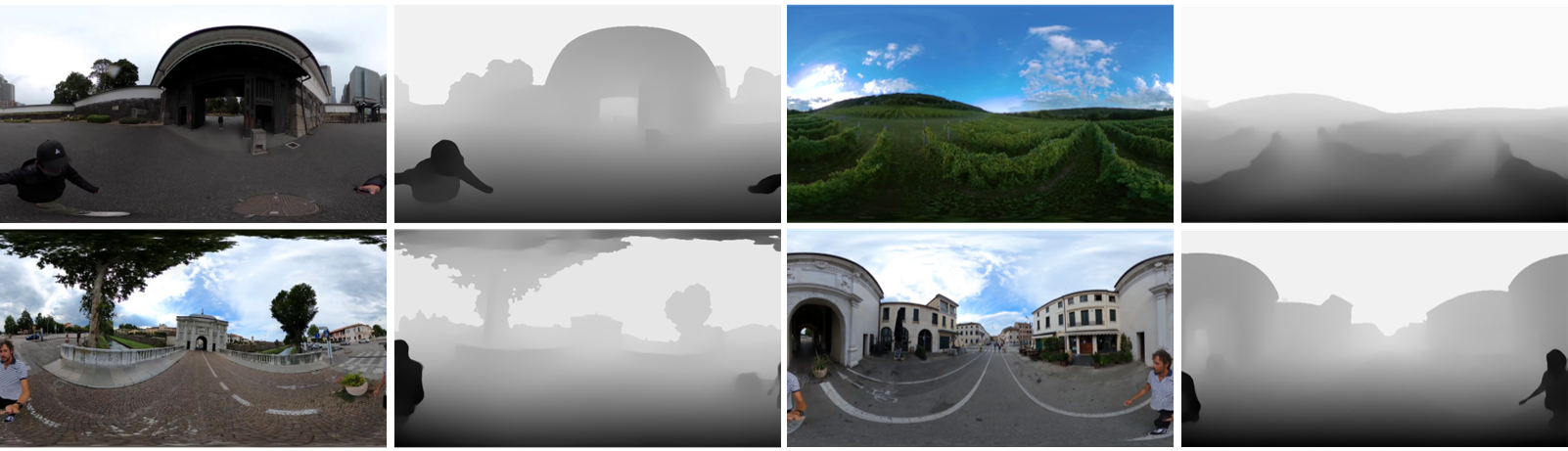}
    \vspace{-18pt} 
  \setlength{\belowcaptionskip}{-10pt}
  \caption{Examples of generated RGB/depth pairs. The color images are video frames acquired from the internet, and the corresponding depth maps is generated through our test-time training method.}
  \label{fig:two}
\end{figure}

We propose the world first generated large-scale dataset \textit{Depth360} that utilizes 360 videos in the wild to solve the unavailability of a general omnidirectional dataset with dense depth information. We first preprocess a video sequence with an SfM and MVS approach to establish quality frame groups that facilitate computing constraints of the sequences. With a horizontal spherical disparity model, we propose novel temporal and geometric consistencies that are unique to 360 videos. By incorporating constraints into test time training through backpropagation, we generate convincing dense depth maps for the corresponding sequence. The generation process is shown in Fig. \ref{fig:one}, and some examples are shown in Fig. \ref{fig:two}.

\subsection{Test-time Training}

\begin{figure*}[ht]
\centering
  \includegraphics[width=0.83\textwidth,keepaspectratio]{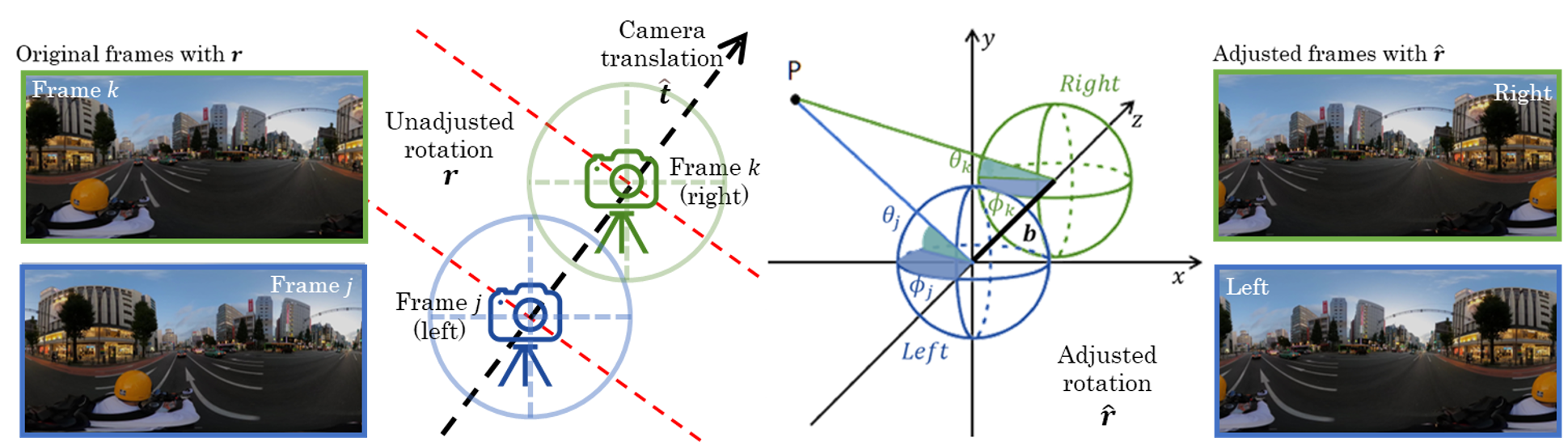}
  \vspace{-8pt} 
  \setlength{\belowcaptionskip}{-15pt}
  \caption{The spatial adjustment process using geometric horizontal spherical disparity models. The left illustration describes original successive frames that satisfy L-R stereo correspondence using the camera poses. The right illustration describes the model to calculate disparity from two frames with a left-right displacement. The $\phi$ denotes the longitude while the $\theta$ denotes the latitude. The $b$ is the baseline acquired from the previous step, and $P$ is the 3D displacement of a target point.
  The left examples show unconstrained frames acquired directly from internet videos, and the right examples show spatially adjusted frames that facilitate the calculation of geometric constraint.}
  \label{fig:four}
\end{figure*}

We propose a test-time training method that first estimates plausible dense estimations utilizing a learning-based prior, and then iteratively fine-tines the parameters during test time with unique constraints established from a certain 360 sequence to generate accurate depth output. Since 360 videos gathered from the internet usually suffer from unconstrained extrinsic and different intrinsic, existing methods often fail to show satisfying performance for dataset creation. On the one hand, depth produced by reconstruction-based methods is usually sparse and erroneous due to distortions. On the other hand, directly applying learning-based methods for frames independently usually results in inconsistent estimation and sub-par accuracy due to the domain gap between perspective and equirectangular formats. With the proposed test-time training method, we take preprocessed video sequences as input and generate geometrically and temporally consistent dense depth maps for each frame. 

We calculate a geometric loss between corresponding frames reprojected from the estimated depth map and stereo pairs’ disparity, in addition to a temporal loss that penalizes the error between flow-based and depth-based projections. In each iteration of fine-tuning a pre-trained depth estimation network, we first generate depth maps for multiple frames with the current network. We then update the parameters according to the calculated geometric and temporal losses to ensure its weight can produce more consistent output for a particular sequence (Fig. \ref{fig:one}).

\textbf{Preprocessing.} 
We exclude dynamic foreground objects from the frames for better calculating camera extrinsic and establish geometric constraints for the respective sequence. Since people are usually the most common dynamic foreground objects in perspective videos \cite{luo2020consistent}, we found this remains true for omnidirectional videos in the wild as well.

We first use OpenVSLAM \cite{sumikura2019openvslam}, an open-source visual SLAM framework, to estimate the pose of the camera $(r,t)$ and the distance $b$ between frame pairs. We then use an off-the-shelf SfM pipeline COLMAP \cite{schonberger2016structure} to acquire sparse depth maps $D^{Recon}$. To improve pose estimate for videos with a strong motion, we apply Mask R-CNN \cite{he2017mask} to obtain static segmentation for more reliable feature point extraction and matching. During this process, we automatically filter out videos with a static viewpoint and vertical motions with estimated poses since they are more challenging in establishing the geometric constraints, and group the remaining videos into consistent short sequences $S$.

\textbf{Spatial adjustment.} 
Since learning-based and reconstruction-based methods are independent of each other and both are scale-invariant, we need to first adjust the scale to match the output before establishing geometric constraints. We achieve this by multiplying all estimated camera translations for a single sequence with a scale factor to match the scale of learning-based depth estimations. For sequence $S_{i}$ with $j$ frames, the scale factor $s_{i}$ is calculated as:
\begin{equation}
S_{i} = {\sum_{j}^{}\frac{D_{j}^{NN}(x)}{D_{j}^{Recon}(x)}/j|D_{j}^{Recon}(x)\neq0}
\end{equation}
where the $D(x)$ is the depth value at pixel $x$ yielded by the learning-based prior before test-time training. The updated camera translation is now $\hat{t}_i=s_{i} \cdot t_{i}$.

While it is usually impossible to create aligned stereo pairs from unconstrained perspective videos due to random camera extrinsic, omnidirectional images have the unique feature of rotation-invariance. For short 360 sequences with minimal vertical movements, we can create aligned left-right stereo image pairs by adjusting the rendering camera rotation to $\hat{r}$ so that the trajectory of frame centers stays parallel to the camera translation $\hat{t}$. This process is demonstrated in Fig. \ref{fig:four}.

\begin{figure*}[ht]
\centering
  \includegraphics[width=0.75\textwidth,keepaspectratio]{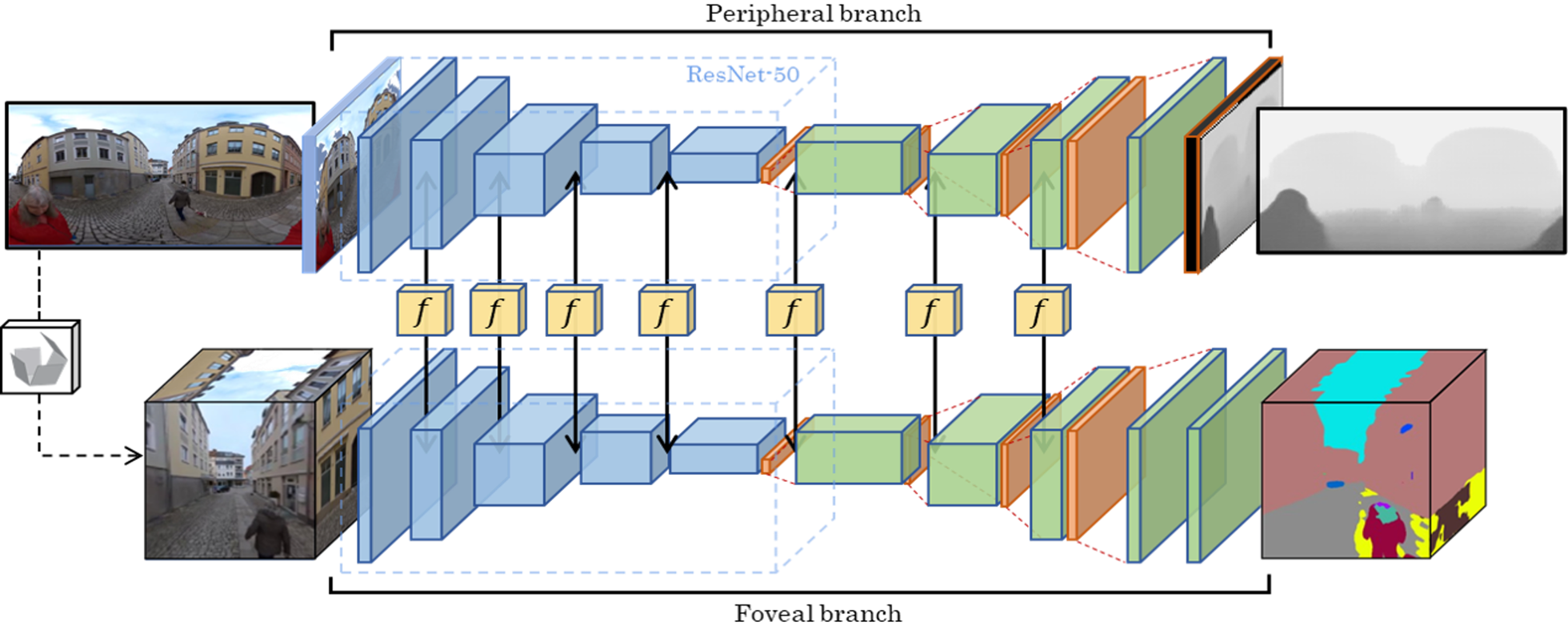}
  \vspace{-8pt} 
  \setlength{\belowcaptionskip}{-15pt}
  \caption{Overview of the proposed end-to-end two-branch multi-task learning network, SegFuse. Structure-wise, the peripheral branch that uses equirectangular projection is capable of capturing global context while the foveal branch uses cubemap projection produces sharper boundaries for local objects. Objective-wise, semantic segmentation and depth estimation are jointly learned to reveal the scene layout and object shapes, while the peripheral branch enforces more consistent depth estimation with a wider FOV, the foveal branch estimating segmentation is more robust to scale changes which frequently appear in a more general dataset. The fusion modules $f$ further facilitate feature sharing between two branches.}
  \label{fig:five}
\end{figure*}

\textbf{Geometric loss.}
To calculate the geometric loss from adjusted left-right image pairs $(j,k)$ with a baseline $b$, we use a modified spherical disparity model from \cite{zioulis2019spherical}. For each point $p$ at $(x,y,z)$ in Cartesian coordinate, we use longitude $\phi$ and latitude $\theta$ in spherical polar coordinate to describe the corresponding point (Fig. \ref{fig:four}). In this sense, the radial distance $r$ to a certain point is $\sqrt{x^2+y^2+z^2}$, and the horizontal disparity is defined as $\delta=(\phi_j-\phi_k,\theta_j-\theta_k)$. Since the baseline $b=(0,0,dz)$ is acquired from the previous step, the disparity is now $\delta=(\frac{\partial\phi}{\partial z},\frac{\partial\theta}{\partial z})$. The transformation between spherical and Cartesian coordinates is omitted to simplify the notations.

To render a target frame $\hat{k}$ from the source frame $j$, each pixel $p=(\phi,\theta)$ on the equirectangular image is a function of the baseline $b$ and the radial distance $r$. Since we already have the generated depth map $D_{j}^{NN}(p)$ for frame $j$, we can compute the target frame $\hat{k}$ with a function: 
\begin{equation}
\hat{k}(p)= \Gamma_{j \rightarrow \hat{k}}(D_{j}^{NN}(p), b_{j \rightarrow k}, j(p))
\end{equation}

Considering the image acquired from online videos are usually not perfect stereo pairs and include dynamic foreground objects, errors often got amplified at certain regions (e.g. top and bottom) on equirectangular projection due to stronger distortion. To alleviate this problem, we further adopt a weight matrix $M(p)=|sin(\phi)||sin(\theta)|$ that assigns different weights for each pixel and aggregates the loss with regard to the distortion level when calculating the geometric loss: 
\begin{equation}
L_{j \rightarrow k}^{geometric}= \sum_{p}^{}||M\hat{k}{p}-Mk(p)||_{2} 
\end{equation}

\textbf{Temporal loss.}
Optical flow is a popular option to check short-term consistency in learning-based video processing for its capability of describing the same scene points in successive frames \cite{ruder2018artistic}. Since depth-estimation networks estimate depth maps independently, the result for a video is usually unstable and inconsistent. To solve the inconsistency between frames of a 360 video, for all frame pairs $(j,k)$ in sequence $S_i$, we further calculate a dense optical flow $f_{j\rightarrow k}$ to ensure a temporal consistency during test-time training. 

It is more suitable to establish short-term and long-term consistency for omnidirectional videos compared to unconstrained perspective videos due to two reasons. First, bad alignment of frames is challenging to cope with for perspective videos while spherical videos can be easily calibrated with simple rotations. Second, while the problem of occlusion remains, objects exiting and re-entering the frame are significantly less prominent in equirectangular videos, making the long-term consistency more reliable. To account for distortions of equirectangular projection, we use a modified version of FlowNet2 \cite{ilg2017flownet}, OmniFlowNet \cite{artizzu2021omniflownet} with a distorted CNN kernel.

For pixel $p=(\phi,\theta)$ on a source equirectangular image $j$, the corresponding pixel $\widetilde{p}$ on the target frame $\widetilde{k}$ is calculated by:
\begin{equation}
\widetilde{p}= p + f_{j \rightarrow k}(p)
\end{equation}
where $f$ denotes the optical flow between two frames. We compute the target frame $\widetilde{k}$ based-on flow with function $F$:
\begin{equation}
\widetilde{k}(p)= F_{j \rightarrow \widetilde{k}}(f_{j \rightarrow k}(p), j(p))
\end{equation}
Similarly, the temporal loss is calculated for each pixel with:
\begin{equation}
L_{j \rightarrow k}^{temporal}= \sum_{p}^{}||\widetilde{k}{p}-k(p)||_{2}
\end{equation}

\textbf{Optimization.}
We then fine-tune the network weights with the combined loss $L_{j \rightarrow k}$ between frame pairs through backpropagation for 10 epochs:
\begin{equation}
L_{j \rightarrow k}= L_{j \rightarrow k}^{geometric}(p) + L_{j \rightarrow k}^{temporal}(p)
\end{equation}
The overall loss is a sum of the geometric loss and the temporal loss calculated over all pixels in video frames, and the network parameters are initialized using a pre-trained network \cite{li2019learning} trained on the Mix 5 dataset \cite{ranftl2019towards}. To reduce the computational cost of computing dense optical flow for image pairs, we calculate the flow between consecutive frames for short-term consistency and left-right pairs for long-term consistency.

\subsection{Implementation Details}
To create the general dataset Depth360, we use the test-time training method to generate convincing depth maps from omnidirectional videos in the wild. We first gathered equirectangular video sequences from the internet that are captured with a hand-held omnidirectional camera. After filtering out samples with strong motion blur, post-editing, and texture-less scenes, we used 30 clips to produce corresponding depth maps. We then fine-tune the weight of the same pre-trained network for each sequence with the geometric and temporal loss using standard backpropagation. By generating consistent depth maps for each sequence with fine-tuned networks after 10 epoch, we create a dataset of paired color images and depth maps with a size of 30,000. Several examples of our generated samples are shown in Fig. \ref{fig:two}. 

\subsection{The Benchmarking Dataset}
To benchmark the effectiveness of the test-time training method and accuracy of the Depth360 dataset, we propose using rendering-based methods to generate a small-scale synthetic dataset via 3D models and virtual cameras. This additional SynDepth360 dataset is motivated by the challenge to directly acquire the ground truth of the internet videos. While the large-scale Depth360 dataset is advantageous to train end-to-end models for single-view depth estimation, the rendered small-scale outdoor 360-degree synthetic dataset with diverse settings is helpful for future research, which we will release together with the Depth360.

\section{SegFuse: Single-view Depth Estimation}

Combining the advantages of a more consistent global context and sharper local details, we propose an end-to-end two-branch multitask learning network called SegFuse. It estimates depth from a single omnidirectional view by mimicking the human eye, as shown in Fig. \ref{fig:five}. In particular, the upper branch regresses depth maps with equirectangular projection, resembling human’s peripheral vision to perceive depth, and the lower branch that estimates semantic segmentation with cubemap projection mimics foveal vision to distinguish between different local objects.

We justify our network design from two aspects. Structure-wise, equirectangular projection is capable of capturing global context but the distortion and a larger FOV restrict its effectiveness against local regions, while cubemap projection produces sharper boundaries for local objects but introduces inconsistency between faces. Objective-wise, since semantic segmentation and depth estimation are two tasks usually jointly learned to reveal the scene layout and object shapes \cite{girshick2015fast} \cite{kim2016unified} \cite{misra2016cross}, while semantic segmentation is more robust to scale changes, we design our two-branch multitasking network that takes advantage of both global context and local details to learn single-view estimation on a more general omnidirectional dataset.

\subsection{Network Structure}
\textbf{The peripheral branch.}
Our peripheral branch regresses a dense global depth estimation from a single view equirectangular image. Its encoder-decoder structure progressively downscales and upscales to the target depth maps. We adopt rectangular filters with changing sizes at the first convolution layer to account for different distortion strengths along the vertical axis of the input equirectangular image. The encoder of this branch shares the same structure of ResNet-50 \cite{he2016deep}, while the decoder consists of four up-projection blocks \cite{laina2016deeper}.

\textbf{The foveal branch.}
Our foveal branch receives reprojected cubemap faces of the input equirectangular image as input and generates semantic segmentation as the output. We choose the semantic segmentation task for the cubemap branch for two reasons. First, although directly regressing depth maps for separate cube maps seems to be a more intuitive choice and has shown some improved performance in similar applications \cite{wang2020bifuse}, the problem of discontinuity at cubemap boundaries is amplified when applied to uncontrolled general samples. We believe that compared to indoor scenes with more uniform and symmetrical structures, our samples generated from online videos are more challenging for the network to learn due to stronger scale ambiguity caused by distinct depth ranges and unsymmetrical depth maps (e.g. sky and ground). This is further verified in our qualitative evaluation.
Second, with undistorted cubemap projection, the foveal branch not only facilitates sharing features of local objects, it can also directly utilize traditional perspective-based model weights to accelerate the learning process, improve the model accuracy, and most importantly, circumvent the challenge of acquiring omnidirectional segmentation ground truth.

Structure-wise, to better facilitate feature fusion at each scale, we set up an identical encoder-decoder network with Resnet-50 encoder and four up-projection modules as the decoder. Instead of incorporating a filter at the first layer to account for distortion, we reproject the equirectangular image to cubemap before feeding it to the first layer. We then incorporate a spherical padding process \cite{wang2020bifuse} to pass feature maps between layers to connect different cube faces. 

\textbf{The fusion scheme.}
To encourage feature sharing between the peripheral branch and the foveal branch, we perform a fusion scheme that lets each branch inform the other with respective feature maps to balance both branches during the training process. Unlike \cite{wang2020bifuse}, we simplify the fusion scheme to improve the training efficiency, and we reduced the number of fused layers to prevent an unstable training process due to different tasks. A more detailed ablation study is presented in the experiment section.

With $m_p$ as the feature map from the peripheral branch and $m_f$ as the feature map from the foveal branch, we first reproject the $m_f$ to $\hat{m_f}$ in equirectangular format and $m_p$ to $\hat{m_p}$ in cubemap projection. We then pass $m_{p}+C(\hat{m_f})$ to the next layer of the peripheral branch and $m_{f}+C(\hat{m_p})$ in the foveal branch respectively. The $C$ denotes a convolution layer. 

\subsection{Loss Functions}
We use supervised loss constraints for both depth estimation and semantic segmentation tasks. For depth estimation, we use inverse Huber loss defined in \cite{laina2016deeper} as the optimizing objective:
\begin{equation}
L^{D}(d)=\begin{cases} |d| & |d| \geq c \\
                           \frac{d^2+c^2}{2c} &  |d|>c 
               \end{cases}
\end{equation}
where $d$ is the difference between the estimated result and the ground truth for each pixel, $c=\max(d)/5$. The loss function $L^S$ for semantic segmentation is a cross-entropy loss between the estimated segmentation $\Bar{S}$ and the result predicted with a pre-train network $S$. Combined, the total loss function can be defined as:
\begin{equation}
L^{total}=L^{D}(D, \Bar{D})+L^{S}(S, \Bar{S})
\end{equation}
During the experiment, segmentation samples are acquired with a pre-trained weight trained on MIT ADE20K dataset \cite{zhou2017scene}. 

\begin{figure}[htb]
\centering
  \includegraphics[width=0.40\textwidth,keepaspectratio]{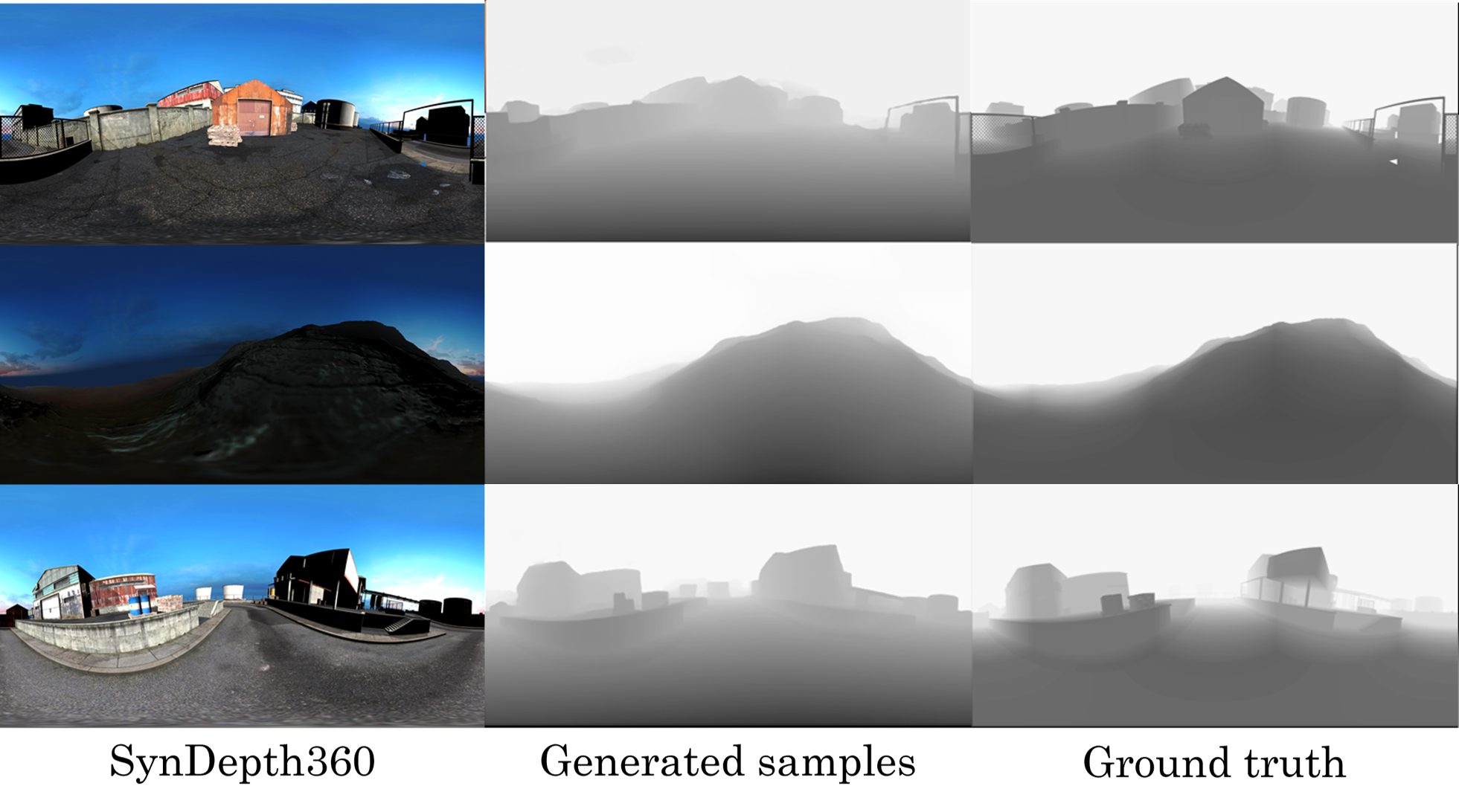}
  \vspace{-10pt} 
  \setlength{\belowcaptionskip}{-10pt}
  \caption{Evaluating the data generation method with the SynDepth360 dataset. The left column is the synthetic samples generated from 3D models, the middle column is the generated data using the test-time training, and the right column is the rendered ground truth.}
  \label{fig:nine}
\end{figure}

\begin{figure}[htb]
\centering
  \includegraphics[width=0.46\textwidth,keepaspectratio]{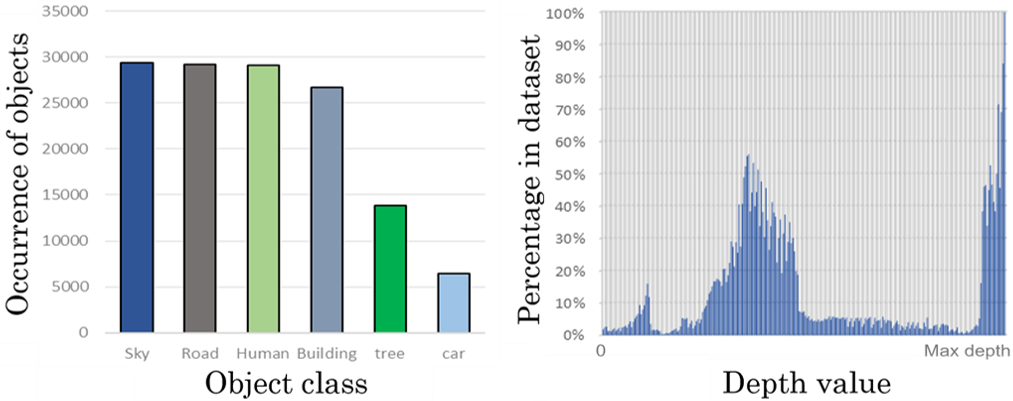}
    \vspace{-8pt} 
  \setlength{\belowcaptionskip}{-5pt}
  \caption{Quantitative evaluations of the dataset. The left figure shows occurrence of top objects in the proposed dataset. The power-law shaped distribution indicates the most predominant background 'sky', 'road' and 'building', while the most occurred foreground objects are 'human' and 'tree'. The right figure shows the distribution of depth values. the leftmost and the rightmost peak manifests that internet videos often use a hand-held capturing fashion with outdoor settings.}
  \label{fig:ten}
\end{figure}

\begin{figure*}[htp]
\centering
  \includegraphics[width=0.87\textwidth,keepaspectratio]{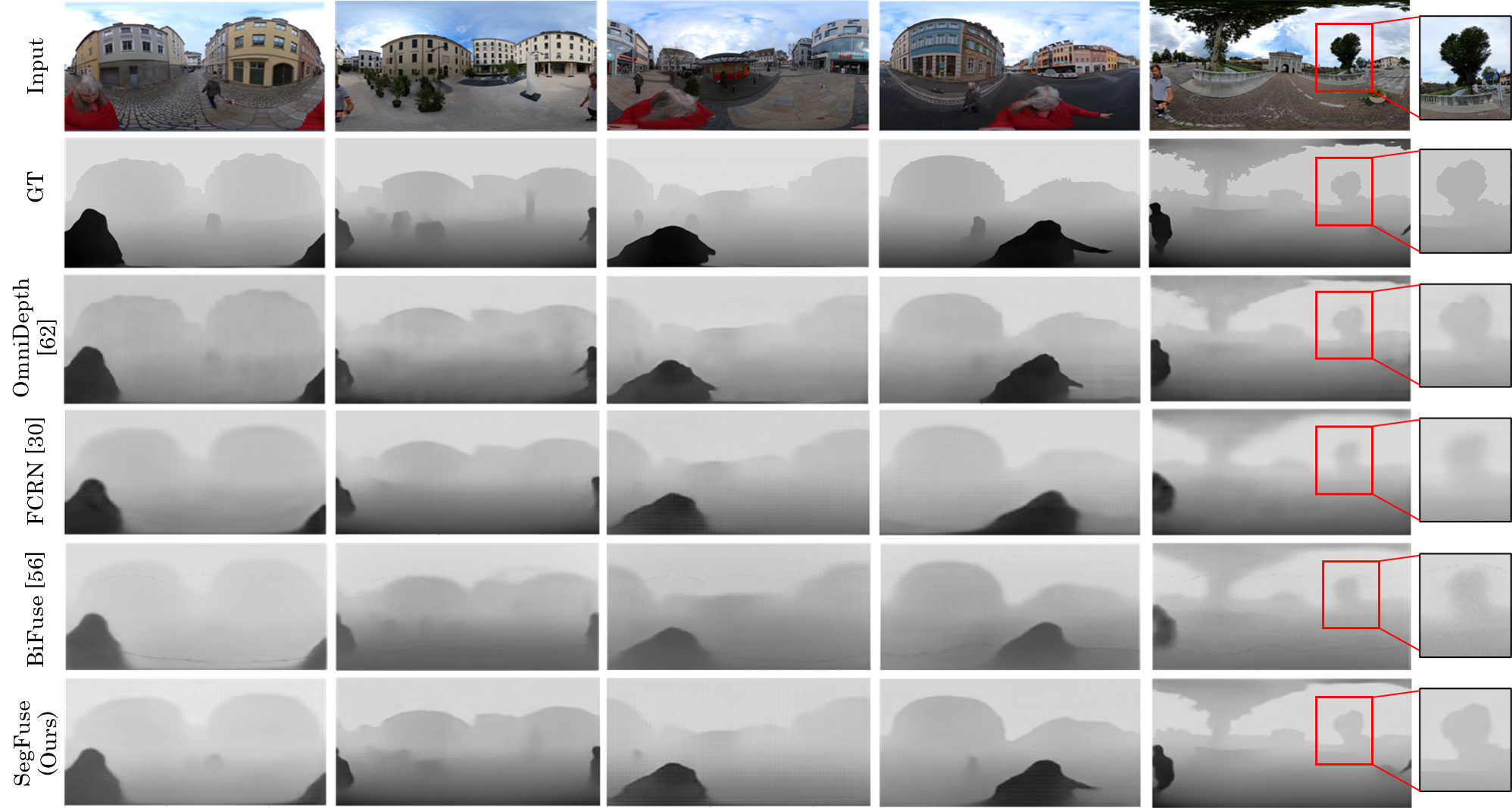}
    \vspace{-8pt} 
  \setlength{\belowcaptionskip}{-10pt}
  \caption{Qualitative comparison with the state-of-the-art methods. Our method generates globally consistent estimation when compared to \cite{zioulis2018omnidepth}, and sharper results at local regions compared to the other methods.}
  \label{fig:six}
\end{figure*}

\section{Experimental Results}
In this section, we first evaluate the quality of the Depth360 dataset against the synthetic dataset SynDepth360 we generated for benchmark purposes. We then evaluate the proposed method against other state-of-the-art single-view depth estimation methods both qualitatively and quantitatively by training on the Depth360 dataset. We further conduct an ablative study to validate the effectiveness of our network design. Both datasets and the source code are available to the community to encourage future research.

\subsection{Dataset Evaluations}

To verify the accuracy of generating depth from internet videos with the proposed test-time training method, we use the synthesized samples from the benchmarking dataset SynDepth360 to evaluate against the ground truth depth maps qualitatively and quantitatively. We further provide a distribution analysis of Depth360 and a quantitative evaluation against existing omnidirectional datasets. 

Qualitatively, samples generated from the synthetic sequences are shown in Fig. \ref{fig:nine}. As most fine details are faithfully reconstructed, and in-scene objects show clear boundaries, we validate that the generated dataset is useful for further single view estimation training. It is worth noting that since the ground truth is rendered with absolute distances with a range of infinity, a slight depth scale discrepancy is presented. Quantitatively, our method achieves the accuracy of 49.5\%, 60.6\%, 70.8\% for $d1$, $d2$, and $d3$
using the metrics for depth prediction from the literature \cite{eigen2015predicting} \cite{godard2017unsupervised}, significantly surpassing naive generation methods. They include omnidirectional models trained with 360 indoor samples, with the highest accuracy of 14.6\%, 22.0\%, 25.5\%, and perspective models trained with mixed samples, with the highest accuracy of 41.7\%, 49.3\%, 59.2\%.

Compared to current state-of-the-art omnidirectional datasets that facilitate single-view depth estimation (Table \ref{tab:dataset}), the proposed dataset achieves higher resolution, larger size, and more diverse outdoor settings. Although the size and quality of model-based datasets (e.g. SceneNet \cite{handa2016scenenet}) can be improved upon using different rendering methods, our dataset maintains the advantages of varied domains and easy extension with a larger video collection. 

\begin{table}[htb]
\centering
\caption{Comparison between state-of-the-art 360 datasets.}
\label{tab:dataset}
\resizebox{0.85\columnwidth}{!}{%
\begin{tabular}{ccccc}
\toprule
    \text{Name}        & \text{Type}       & \text{setting}       & \text{resolution}       & \text{\# images}     \\
     \midrule
 PanoSunCG \cite{wang2018self} & synthetic & indoor & 0.13Mpx & 25000 \\
 SceneNet \cite{handa2016scenenet} & synthetic & indoor & 0.13Mpx & 25000* \\
 Stanford 2D-3D \cite{armeni2017joint} & real & indoor & 0.13Mpx & 25000* \\
 Matterport3D \cite{chang2017matterport3d} & real & indoor & 0.13Mpx & 25000* \\
 Proposed dataset & \textbf{real} & \textbf{outdoor} & \textbf{1.03Mpx} & \textbf{30000}\\
 \bottomrule
\end{tabular}}
\end{table}

From the depth value distribution analysis (Fig. \ref{fig:ten}, right) of the Depth360 dataset, we can observe three major peaks. The leftmost peak manifests the hand-held 360 camera user, while the rightmost peak shows a common sky background. The normal distribution in the middle presents other objects in the scene such as buildings and trees. This can be validated through the occurrence of top objects (Fig. \ref{fig:ten}, left) in the proposed dataset. From the power-law shaped distribution, we can observe that the most predominant background objects are 'sky', 'road' and 'building', and most occurred foreground objects are 'human'.  

\subsection{SegFuse Evaluations}
\subsubsection{Implementation Details}
We implement the SegFuse network with the Pytorch framework \cite{paszke2017automatic} 1.4 and train models with a configuration of Nvidia RTX 2080Ti GPU, i7-7800X CPU, and 32GB RAM. We randomly split samples into training and validation datasets from the dataset with a ratio of 90\% and 10\%. During training, we use Adam optimizer \cite{kingma2014adam}, a learning rate of 3e-4 and, a batch size of 1. The peripheral branch uses Xavier initialization \cite{glorot2010understanding} while the foveal branch initializes with ImageNet pretrained weights. The same metrics from the literature are used for quantitative evaluation \cite{godard2017unsupervised}. Our current implementation takes smaller batches due to graphics memory restrictions. We expect a more stable training process with a better hardware configuration, with potential improvements such as batch normalization. Our method costs approximately 100ms with the same configuration to predict a single equirectangular image, favoring interactive frame-rate for applications.

\begin{table*}[t]
\label{tab:accuracy}
\begin{center}
\resizebox{0.80\linewidth}{!}{\begin{minipage}{\textwidth}
\caption{Quantitative results against other single-view omnidirectional depth estimation methods.}
\begin{tabular*}{\textwidth}{c @{\extracolsep{\fill}} ccccccc}
\toprule
Method & Abs Rel $\downarrow$ & Sq Rel $\downarrow$ & RMSE $\downarrow$ & RMSE log $\downarrow$ & $\delta < 1.25$ $\uparrow$ & $\delta < 1.25^2$ $\uparrow$ & $\delta < 1.25^3$ $\uparrow$ \\
\midrule
\multicolumn{8}{c}{Real domain (all models are trained with Depth360 dataset)} \\
\midrule
OmniDepth \cite{zioulis2018omnidepth} & 0.3375 & 0.1967 & 4.3049 & 0.7836 & 80.16\% & 89.78\% & 91.93\%\\
BiFuse	\cite{wang2020bifuse} & 0.3596 & 0.8615 & 5.0725 & 0.8316 & 40.13\% & 59.17\% & 67.92\%\\
FCRN \cite{laina2016deeper} & 0.2384 & 0.4057 & 4.8599 & 0.7839 & 80.59\% & 90.42\% & 93.88\%\\
SegFuse (Ours) & \textbf{0.2275} & \textbf{0.1588} & \textbf{4.0442} & \textbf{0.7777} & \textbf{82.26\%} & \textbf{91.35\%} & \textbf{94.22\%}\\
\midrule
\multicolumn{8}{c}{Synthetic domain (all models are trained with SynDepth360 dataset)} \\
\midrule
OmniDepth \cite{zioulis2018omnidepth} & 0.1171 & 0.1753 & 0.3819 & 0.0844 & 91.30\% & 94.04\% & 96.35\%\\
BiFuse	\cite{wang2020bifuse} & 0.1473 & 0.1978 & 0.4619 & 0.1012 & 75.12\% & 81.77\% & 84.69\%\\
FCRN \cite{laina2016deeper} & 0.1017 & 0.1525 & 0.3771 & 0.0776 & 93.48\% & 95.89\% & 97.79\%\\
SegFuse (Ours) & \textbf{0.0973} & \textbf{0.1510} & \textbf{0.3209} & \textbf{0.0734} & \textbf{94.74\%} & \textbf{96.61\%} & \textbf{98.03\%}\\
\bottomrule
\end{tabular*}
\vspace{-25pt}%
\end{minipage}}
\end{center}
\end{table*}%

\subsubsection{Qualitative Results}

We present qualitative results of single-view depth estimation from omnidirectional images with different methods including Omnidepth \cite{zioulis2018omnidepth} and FCRN \cite{laina2016deeper}. As we can observe in Fig. \ref{fig:six}, when tested on unseen equirectangular images with challenging outdoor settings, our method generates better sharper estimation while maintaining a smooth global depth map prediction. This can be attributed to the foveal branch that improves local details. More qualitative results with diverse settings are included in the supplementary material.

\begin{figure}[htb]
\centering
  \includegraphics[width=0.42\textwidth,keepaspectratio]{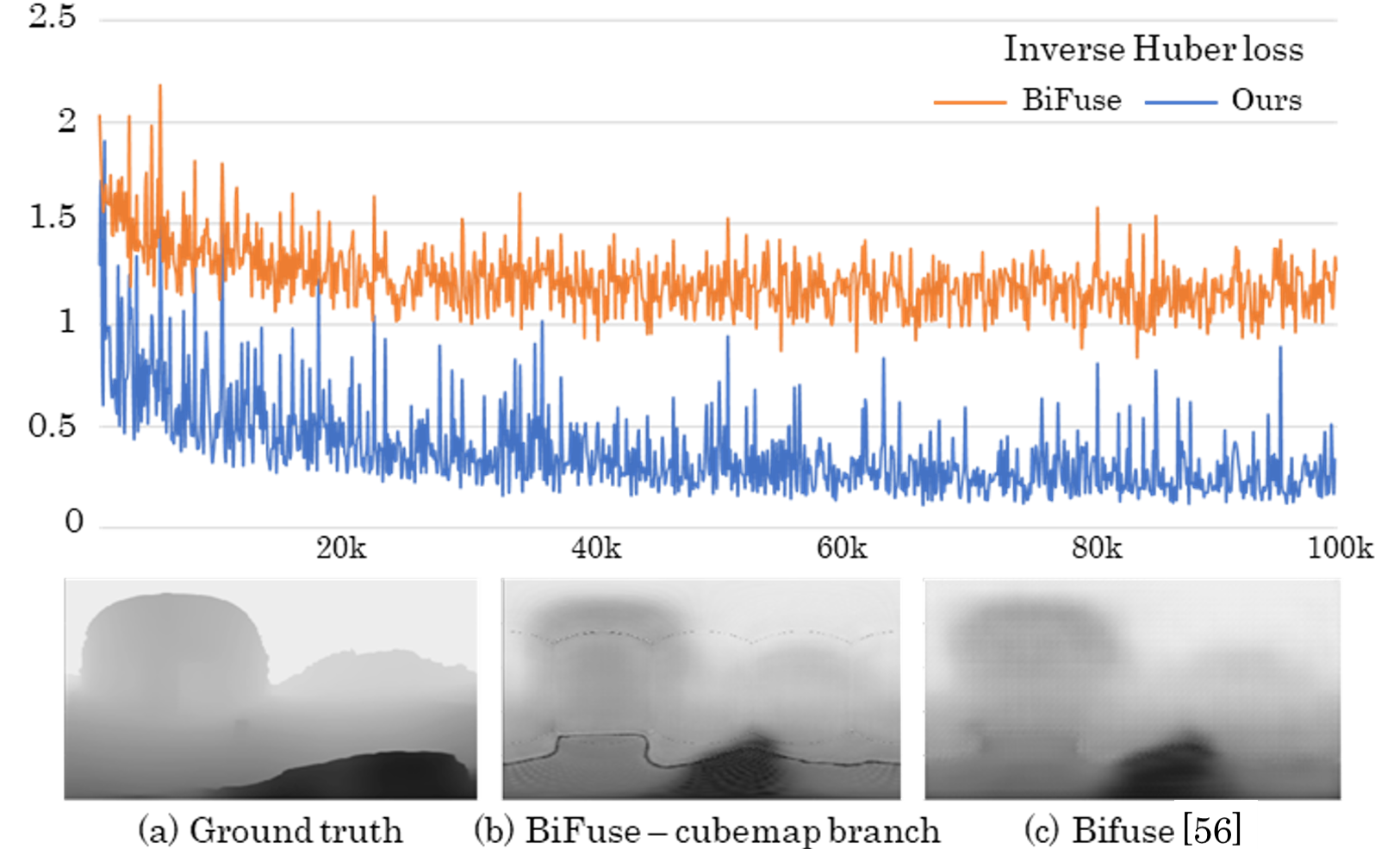}
  \vspace{-5pt} 
  \setlength{\belowcaptionskip}{-10pt}
  \caption{To validate our network design, we evaluate against BiFuse \cite{wang2020bifuse}, a cubemap-based depth fusion method. When trained with outdoor samples, SegFuse converges much faster (the blue line) while the cubemap-based depth regression struggles to effectively merge faces and learn outdoor settings (the orange line). (b) shows the result of regressed depth from the cubemap branch of BiFuse, and (c) shows the final fused output of BiFuse.}
  \label{fig:seven}
\end{figure}

As we argued that for challenging outdoor cases with wildly changing ranges and unsymmetrical distributions, directly using cubemap projection to regress depth maps for each face and fusing with equirectangular estimation afterward shows sub-optimal performance. Such inconsistencies at face boundaries are presented in Fig. \ref{fig:six} (BiFuse \cite{wang2020bifuse}). We offer a detailed convergence analysis of the proposed method against BiFuse that uses cubemap projection to fuse depth for outdoor samples. The results can be observed in Fig. \ref{fig:seven}. We compare the performance via inverse Huber loss when both networks are trained with the Depth360 dataset. We show that our method converges much faster (the blue line) with the help of latent information shared by the pretrained semantic segmentation weight, while the cubemap-based depth regression struggles to effectively merge faces and learn outdoor settings (the orange line). As we can see in the bottom half of Fig. \ref{fig:seven}, the middle figure shows the result of regressed depth from the cubemap branch of BiFuse, and the right figure shows the final fused output of BiFuse. Clear boundaries between faces 
result in deteriorated fused output when compared to a single-branch architecture such as FCRN. 

\subsubsection{Quantitative Results}
Adopting the metrics for depth prediction from the literature \cite{eigen2015predicting} \cite{godard2017unsupervised}, Table 2 presents the quantitative evaluation of our method against the state-of-the-art single-view omnidirectional depth estimation methods in both real-world and synthetic domains. We can observe that SegFuse successfully captures the features of the outdoor dataset when compared to other methods. Overall, our method shows favorable results against FCRN, Omnidepth, and BiFuse. We further evaluate the performance in indoor settings by training networks with 3D60 dataset \cite{zioulis2018omnidepth}, which consists of SunCG \cite{wang2018self}, SceneNet \cite{handa2016scenenet}, Stanford2D3D \cite{armeni2017joint}, Matterport3D \cite{chang2017matterport3d}. We benchmark against the ground truth depth with filled-in values for invalid pixels like FCRN \cite{laina2016deeper}. Table 3 shows a comparable accuracy of SegFuse with the state-of-the-art designed for indoor predictions \cite{wang2020bifuse}, and better performance against other omnidirectional methods.

\begin{table}[htb]
\centering

\caption{Qualitative results of indoor-only settings.}
\resizebox{0.9\columnwidth}{!}{%
\begin{tabular}{c*{5}{>{}c<{}}}
 \toprule
  Method & RMSE & RMSE log & $\delta < 1.25$ & $\delta < 1.25^2$ & $\delta < 1.25^3$\\
 \midrule
    OmniDepth \cite{zioulis2018omnidepth} & 0.6364 & 0.1358 & 77.30\% & 91.24\% & 97.21\%\\
    BiFuse	\cite{wang2020bifuse} & \textbf{0.5639} & 0.1007 & \textbf{85.12\%} & 93.38\% & \textbf{98.16\%}\\
    FCRN \cite{laina2016deeper} & 0.6429 & 0.1286 & 78.08\% & 92.09\% & 97.33\%\\
    SegFuse (Ours) & 0.5729 & \textbf{0.0986} & 84.38\% & \textbf{94.34\%} & 98.07\%\\
 \bottomrule
\end{tabular}}

\vspace{0.2cm}

\caption{Ablation results of the foveal branch.}
\resizebox{0.90\columnwidth}{!}{%
\begin{tabular}{c*{5}{>{}c<{}}}
 \toprule
  Method & RMSE & RMSE log & $\delta < 1.25$ & $\delta < 1.25^2$ & $\delta < 1.25^3$\\
 \midrule
Peripheral only & 4.9281 & 0.8979 & 57.79\% & 74.20\% & 78.11\%\\
 SegFuse & \textbf{4.0442} & \textbf{0.7777} & \textbf{82.26\%} & \textbf{91.35\%} & \textbf{94.22\%}\\ 
 \bottomrule
\end{tabular}}

\vspace{0.2cm}

\caption{Ablation results of connected layers.}
\resizebox{0.72\columnwidth}{!}{%
\begin{tabular}{c*{5}{>{}c<{}}}
\toprule
    & \text{RMSE}        & \text{RMSE log}       & \text{$\delta < 1.25$}       & \text{$\delta < 1.25^2$}       & \text{$\delta < 1.25^3$}     \\
     \midrule
 1 & 4.9297 & 0.8933 & 70.13\% & 79.17\% & 87.92\%\\
 2 & 4.3782 & 0.8126 & 77.85\% & 86.83\% & 92.50\%\\
 3 & 4.0442 & \textbf{0.7777} & \textbf{82.26\%} & 91.35\% & \textbf{94.22\%}\\ 
 4 & \textbf{4.0168} & 0.7994 & 81.67\% & \textbf{91.75\%} & 93.82\%\\
 \bottomrule
\end{tabular}}
\vspace{-10pt}%
\end{table}

\subsubsection{Ablation Studies}
Finally, we perform an ablation analysis between the SegFuse and learning without the foveal branch. We use the same training settings with and without fusing the foveal branch with the peripheral branch, and the quantitative evaluation is shown in Table 4. In addition to better accuracy, we also find that the converging speed when training with SegFuse is almost 2x faster at the beginning thanks to the pre-trained segmentation weight. This shows the additional benefit of using a multi-task architecture to solve the depth estimation problem. We then compare the accuracy of the SegFuse network with different numbers of fused layers at the decoder. We find that while connecting three and four layers both achieve close performance, using three fusion blocks usually provides a slightly more stable training process and improved efficiency. A quantitative ablation study is presented in Table 5.

\subsection{Depth-based Applications}
High-quality depth estimation from a single 360 image enables a wide range of interesting applications. We take visual effects as an example to showcase the strength of our method in virtual reality. We first use the proposed method to estimate a high-quality dense depth map from an input omnidirectional RGB image. We then project per-pixel depth values onto a 3D sphere to render a pseudo-reconstructed scene with mesh. This facilitates augmenting the original scenes with effects such as volumetric snowing and flooding. A preview is shown in Fig. \ref{fig:eleven}, and full video results are included in the supplementary material.

\begin{figure}[htb]
\centering
  \includegraphics[width=0.43\textwidth,keepaspectratio]{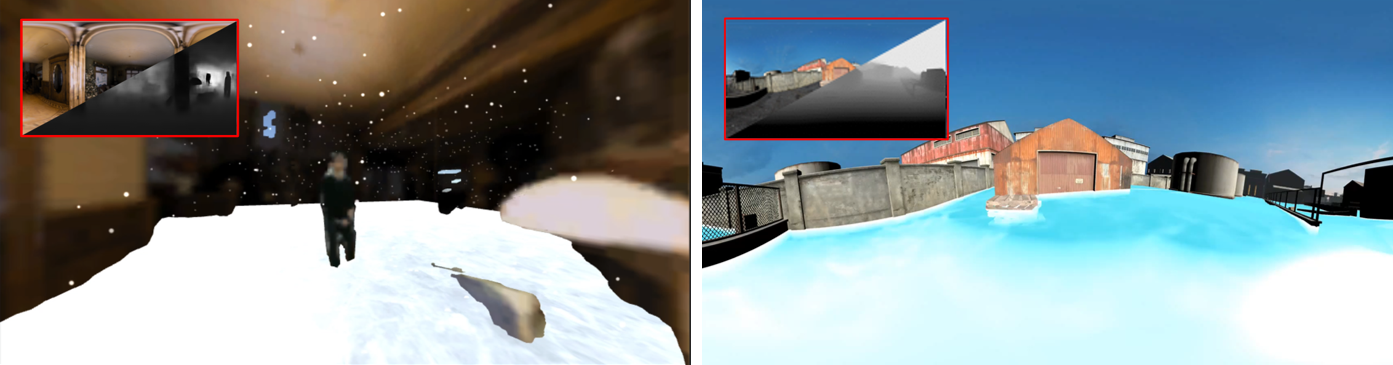}
  \vspace{-8pt} 
  \setlength{\belowcaptionskip}{-12pt}
  \caption{An example of depth-enabled applications. By estimating corresponding depth maps from an input 360 image, we add volumetric effects to the scene such as snowing (left) and flooding (right).}
  \label{fig:eleven}
\end{figure}

\section{Failure Cases and Future Work}
Although our data generation method can be applied to larger-scale collections to extend the size of datasets, it shows several limitations. First, online omnidirectional videos present unbalanced distributions, favoring specific scenarios (e.g. urban street views). Second, when establishing the baseline, SfM and MVS methods show sub-optimal results when there are texture-less surfaces or reflective materials in the scene. Scenes with excessive dynamic foreground objects or strong motions are problematic for a pseudo-stereo system to acquire accurate geometric consistency. Future work could alleviate these problems by adopting improved SfM algorithms and scaling to a larger variety of input collections. For depth estimation, the current implementation only accepts smaller batch sizes due to hardware limitations. We expect to improve the efficiency of the network and enable more stable training with better normalization methods. 
\section{Conclusion}
In this paper, we first propose to utilize the unlimited source of data, 360 videos from the internet, to overcome the scarcity of a general omnidirectional dataset. We propose geometric and temporal constraints that are unique to 360 videos and use test-time training to generate high-quality depth maps. To fully benefit from our dataset, Depth360, we propose an end-to-end two-branch multitask network, SegFuse, that mimics human vision to estimate depth from a single omnidirectional image. With peripheral vision to perceive the depth of the scene and foveal vision to distinguish between objects, our network shows favorable results against state-of-the-art methods.

\acknowledgments{
This research was supported by JST-Mirai Program (JPMJMI19B2), JSPS KAKENHI (19H01129, 19H04137, 21H0504) and the Royal Society (IES\textbackslash R2\textbackslash 181024).}

\bibliographystyle{abbrv-doi}
\bibliography{template}
\end{document}